\documentclass[runningheads]{llncs}
\usepackage[T1]{fontenc}
\usepackage{graphicx}
\usepackage{amsfonts}
\usepackage{booktabs}
\usepackage{amsmath}
\usepackage{array}
\usepackage{adjustbox}
\usepackage{hyperref}

\begin{document}

\title{DAMamba-UNet3D: A Parameter-Efficient Mamba State Space U-Net with Dynamic Adaptive Scan for 3D Medical Image Segmentation}
\titlerunning{Parameter-Efficient DAMamba-UNet3D}

\author{Mohammad Arafat Hussain, Ellen Grant, Yangming Ou}
\authorrunning{Hussain et al.}
\institute{Boston Children's Hospital, Harvard Medical School \\
    \email{mohammad.hussain@childrens.harvard.edu}}

\maketitle

\begin{abstract}
We propose parameter-efficient SSM-based U-Net architectures for 3D medical image segmentation. Convolutional U-Nets afford $O(n)$ local mixing per layer but lack explicit global context; transformers provide global reasoning at $O(n^2)$ cost in sequence length $n$. State-space models (SSMs), such as Mamba, offer $O(n)$ global propagation per block. Yet, existing medical SSM segmenters rely on fixed scan patterns and large parameter budgets. Dynamic Adaptive Scan (DAS), which learns data-dependent reordering before selective scan, has not been applied to medical imaging or extended to 3D volumes. We propose \textbf{DAMamba-UNet3D}, a hybrid encoder--decoder that integrates tri-plane 3D-DAS blocks at encoder stages E2--E4 while retaining convolutions elsewhere (${\sim}5.3$\,M parameters). On BraTS~2020 five-fold cross-validation, DAMamba-UNet3D achieves mean Dice $0.815 \pm 0.013$ (full-volume per-case evaluation) at ${\sim}13\times$ lower parameter cost than SegMamba ($0.824 \pm 0.014$, ${\sim}70$\,M). At comparable scale, \textbf{DAMamba-L} (${\sim}70$\,M), a wide DAS-native variant with encoder-only DAMamba and a convolutional bottleneck, reaches $0.829 \pm 0.012$, surpassing retrained SegMamba by $0.5$\,pt. Component ablations show that encoder-only DAS placement is critical as bottleneck and decoder SSM blocks lower Dice. Together, the results suggest that learned tri-plane DAS in a hybrid U-Net is competitive with, and under our large-scale design may improve upon, SegMamba's fixed Tri-orientated Mamba (ToM) scanning on BraTS~2020. Code: \href{https://github.com/marafathussain/DAMamba-UNet3D}{https://github.com/marafathussain/DAMamba-UNet3D}.


\end{abstract}

\section{Introduction}
\label{sec:introduction}
Three-dimensional (3D) volumetric segmentation requires both precise local boundary delineation and coherent global spatial context~\cite{isensee2021nnu}, under tight compute and parameter budgets. For a volume with $n$ voxels (or tokens), convolutional, transformer, and state-space architectures trade accuracy against complexity differently: \emph{Convolutional U-Nets}~\cite{ronneberger2015unet} are parameter-efficient with $O(n)$ compute per layer for local mixing, but each layer aggregates information only within a limited receptive field; global 3D context requires depth scaling rather than explicit long-range mixing in one pass. \emph{Vision transformers}~\cite{wang2021transbts,cao2023swin} provide global context via self-attention at $O(n^2)$ memory and compute. For example, a $128^3$ patch contains $n \approx 2.1$\,M tokens, prohibitive for naive global attention on a single clinical GPU. \emph{State-space models (SSMs)}, especially Mamba~\cite{gu2023mamba}, propagate information along a 1D sequence in $O(n)$ time and memory per block, offering global context without quadratic attention maps.

SegMamba~\cite{ma2024segmamba} is the state-of-the-art representative of this third family for dense 3D segmentation. It embeds Mamba blocks in a U-shaped 3D network with Tri-orientated Mamba (ToM) modules that serialize features along three anatomical directions at $O(n)$ cost per scan. At ${\sim}70$\,M parameters, however, SegMamba's scan trajectories are fixed by design; the model achieves long-range context through capacity and multi-orientation flattening rather than content-adaptive reordering.

Recently, DAMamba~\cite{damamba2025} introduced \emph{Dynamic Adaptive Scan} (DAS) for 2D natural-image recognition. DAS is learnable, input-dependent spatial reordering via deformable resampling before selective scan. To our knowledge, DAS has not been used in medical imaging, nor extended beyond 2D. We address both gaps with DAMamba-UNet3D, the first DAS-based architecture for medical image segmentation, with a tri-plane formulation that lifts DAS to 3D volumes.

Our contributions are: \textbf{1. First DAS application to medical segmentation.} We transfer DAS from 2D natural-image classification to dense 3D medical segmentation for the first time, embedding it in a hybrid U-Net. \textbf{2. 2D-to-3D DAS extension.} We lift the 2D DAS operator to volumes via tri-plane adaptive scanning (axial, coronal, sagittal) fused before volumetric selective scan, as an alternative to hand-designed multi-orientation flattening. \textbf{3. Parameter-efficient DAMamba-UNet3D.} Convolutions in the stem, bottleneck, and decoder for local detail; tri-plane DAS--SSM blocks only at encoder stages (about 5.3\,M parameters, ${\sim}13\times$ fewer than SegMamba). \textbf{4. Two-scale validation on BraTS~2020.} The compact model reaches near-parity with retrained SegMamba ($0.815$ vs.\ $0.824$ mean Dice); DAMamba-L (${\sim}70$\,M), with stacked encoder DAMamba and a convolutional bottleneck, surpasses SegMamba ($0.829$ vs.\ $0.824$), suggesting that learned DAS in an encoder-only hybrid design can outperform fixed ToM under identical training and evaluation protocol.

\section{Methodology}
\label{sec:methodology}

\subsection{Data and Preprocessing}
\label{sec:data}
We use the BraTS~2020 training set~\cite{mehta2022qu} as our evaluation benchmark consisting of 369 subjects, four modalities (T1, T1ce, T2, FLAIR), and labels for ET, TC, and WT. Necrosis (label~4) is remapped to~3. Each modality is z-scored over nonzero brain voxels. \textbf{Training.} Random $128^3$ patches are sampled each iteration. \textbf{Augmentation.} We used random flips, intensity shift/scale ($\pm0.1$), and Gaussian noise ($\sigma{=}0.01$). \textbf{Cross-validation.} We used five subject-level folds.

\subsection{State-Space Model and Selective Scan}
\label{sec:ssm}

A state-space model (SSM) maps an input $x(t)$ to an output $y(t)$ via a latent state $h(t) \in \mathbb{R}^{d_{\mathrm{state}}}$:
\begin{equation}
h'(t) = A\, h(t) + B\, x(t), \qquad y(t) = C\, h(t),
\end{equation}
with $A \in \mathbb{R}^{d_{\mathrm{state}} \times d_{\mathrm{state}}}$, $B \in \mathbb{R}^{d_{\mathrm{state}} \times 1}$, and $C \in \mathbb{R}^{1 \times d_{\mathrm{state}}}$. Discretizing with step size $\Delta$ yields
\begin{equation}
h_t = \bar{A}\, h_{t-1} + \bar{B}\, x_t, \qquad y_t = C\, h_t,
\end{equation}
where $\bar{A} = e^{\Delta A}$ and $\bar{B} = (\Delta A)^{-1}(e^{\Delta A} - I)\,\Delta B$.

In our network, $\mathbf{X} \in \mathbb{R}^{C \times D \times H \times W}$ denotes a 3D feature map at a U-Net stage (batch dimension omitted). After DAS reordering (Sec.~\ref{sec:dassm3d}), its $N = D \cdot H \cdot W$ voxels form a 1D sequence; voxel $t$ supplies the token $x_t$ fed to Eq.~(2), and $h_t$ stores context accumulated along that scan path.

\textbf{Selective scan (Mamba).} Classical SSMs keep $A$, $B$, $C$, and $\Delta$ fixed. Mamba~\cite{gu2023mamba} instead predicts input-dependent $\Delta_t$, $B_t$, and $C_t$ from each $x_t$, gating what is carried forward. Applied to the voxel sequence above, this yields long-range mixing in $O(N)$ time per block. We use $d_{\mathrm{state}}{=}16$, the standard setting in Mamba~\cite{gu2023mamba} and DAMamba~\cite{damamba2025}.

\subsection{Dynamic Adaptive Scan and Tri-Plane DAMamba Block}
\label{sec:dassm3d}

Each \textbf{DAMamba block} (Fig.~\ref{fig:architecture}) combines Dynamic Adaptive Scan (DAS)~\cite{damamba2025} with the selective scan of Sec.~\ref{sec:ssm}. We adopt the 2D DAS operator from~\cite{damamba2025} and extend it to volumes via tri-plane scanning (step~2 below); our contribution is this medical-first use and 3D formulation, not a redesign of the core DAS mechanism.

\textbf{Dynamic Adaptive Scan (DAS).} Unlike fixed scans (row-major, zigzag, or multi-orientation flattening in prior Mamba segmenters), DAS learns data-dependent sampling~\cite{damamba2025}. An offset prediction network (OPN) predicts 2D offsets $\Delta p \in \mathbb{R}^{H \times W \times 2}$ from the feature map. Predicted positions $p'[h,w] = p[h,w] + \Delta p[h,w]$ determine sampling locations; features are gathered via bilinear interpolation $\phi$:
\begin{equation}
X'[h,w] = \phi(p'[h,w], X), \quad \phi(\mathbf{a}, X) = \sum_{r_x,r_y} g(a_x, r_x)\, g(a_y, r_y)\, X[r_y, r_x, :],
\end{equation}
where $\mathbf{a} = (a_x, a_y)$ is the sampling position (e.g.\ $p'[h,w]$), $X$ is the input feature map, $(r_x, r_y)$ index the four nearest grid neighbors, and $g(\cdot,\cdot)$ is the 1D linear interpolation weight. The resampled features $X'$ define an input-adaptive scan order before SSM processing, allowing the model to emphasize informative regions (e.g., tumor boundaries). For 3D features $\mathbf{X} \in \mathbb{R}^{B \times C \times D \times H \times W}$, each block proceeds as follows: 

\begin{enumerate}
    \item Pointwise expansion (channel factor~2), depthwise $4^3$ convolution, and SiLU activation.
    \item \textbf{Tri-plane DAS.} The 2D DAS operator is applied on axial ($H{\times}W$), coronal ($D{\times}W$), and sagittal ($D{\times}H$) planes and averaged:
    \begin{equation}
    \mathrm{DAS}_{3\mathrm{D}}(\mathbf{X}) = \tfrac{1}{3}\big(\mathrm{DAS}_{\mathrm{ax}} + \mathrm{DAS}_{\mathrm{cor}} + \mathrm{DAS}_{\mathrm{sag}}\big).
    \end{equation}
    \item \textbf{Selective scan.} The reordered map is processed as in Sec.~\ref{sec:ssm}.
    \item Normalization, projection, and a residual connection. The block thus adds global context at $O(N)$ cost without attention maps that scale as $N^2$.
\end{enumerate}

\subsection{DAMamba-UNet3D Architecture}
\label{sec:architecture}

\begin{figure}[t]
\centering
\includegraphics[width=\textwidth]{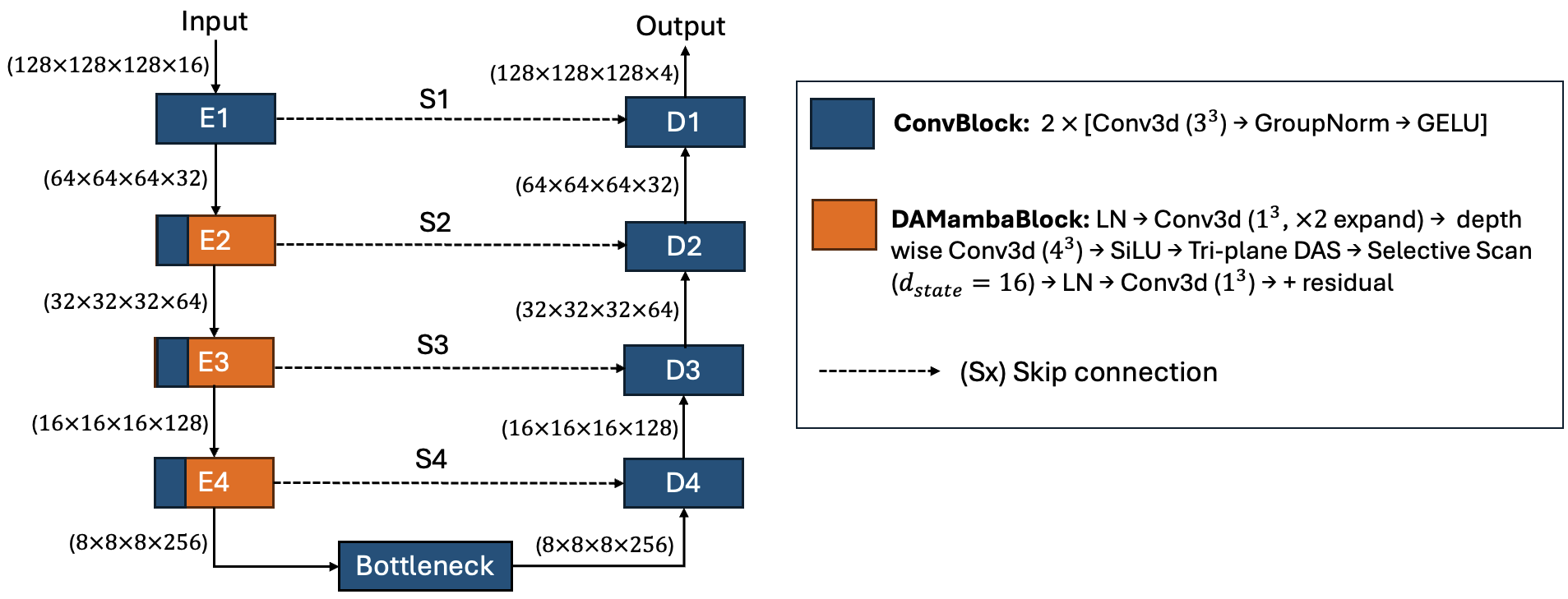}
\caption{DAMamba-UNet3D block diagram with stage-wise feature dimensions ($D{\times}H{\times}W{\times}C$; example shown for $128^3$ training patches). Blue: convolutional blocks (E1, decoder); orange: encoder stages E2--E4, each convolutional block followed by a tri-plane DAMamba block.}
\label{fig:architecture}
\end{figure}

DAMamba-UNet3D is a 4-level 3D U-Net with base width $C{=}16$ (channels $C, 2C, 4C, 8C, 16C$ at encoder stages). The stem (E1), bottleneck, and entire decoder are convolutional. At each encoder stage E2--E4, a convolutional block (two $3^3$ convolutions) extracts local features, followed by one tri-plane DAMamba block (conv $\rightarrow$ DAS $\rightarrow$ selective scan; Sec.~\ref{sec:dassm3d}) for global context. Fig.~\ref{fig:architecture} summarizes the encoder--decoder layout and feature dimensions in $D{\times}H{\times}W{\times}C$ format for $128^3$ training patches (four input modalities, four output channels). 

\textbf{DAMamba-L (large-scale variant).} To evaluate DAS at SegMamba-scale capacity, we train DAMamba-L (${\sim}70$\,M), a wide DAS-native U-Net with SegMamba-aligned stage widths, two stacked DAMamba blocks per encoder stage (E2--E4), and a convolutional bottleneck (no SSM at the bottleneck), mirroring the encoder-only placement validated on the compact model (Sec.~\ref{sec:ablation}).

\subsection{Loss and Training}
\label{sec:loss_training}

We train our DAMamba-UNet3D with the modulated combo loss as~\cite{taghanaki2019combo,hosseini2025pmdice}:
\begin{equation}
\mathcal{L} = (\mathcal{L}_{\mathrm{CE}} + \delta) \cdot (\mathcal{L}_{\mathrm{Dice}}^{\mathrm{mod}} + \delta), \quad \delta{=}0.1,\; \gamma{=}2,
\end{equation}
where modulation weights $w_i^{(c)} = |p_i^{(c)} - y_i^{(c)}|^\gamma$ focus learning on hard voxels and boundaries. \textbf{Optimization.} We used AdamW optimizer, learning rate of $5{\times}10^{-5}$, cosine decay of $10^{-6}$, weight decay of $10^{-4}$, and gradient clipping at~1.0, 50 epochs. 

\subsection{Evaluation Protocol}
\label{sec:eval}

All results in this paper use the same full-volume evaluation pipeline. For each fold's best training checkpoint, we reconstruct the complete BraTS volume with sliding-window inference (patch $128^3$, overlap~0.5, Gaussian blending). We then compute per-case Dice and HD95 (mm) for ET, TC, and WT under the BraTS-official empty-region convention (both empty $\rightarrow$ excluded; empty target with non-empty prediction $\rightarrow$ Dice~$=0$). Reported mean Dice averages ET, TC, and WT per case, then aggregates across cases and folds. This protocol avoids optimistic estimates from slice batches or cropped validation regions.

\section{Results}
\label{sec:results}

All experiments in this paper are evaluated on BraTS~2020 (Sec.~\ref{sec:data}). Segmentation performance by DAMamba-UNet3D (${\sim}5.3$\,M), SegMamba~\cite{ma2024segmamba} (${\sim}70$\,M), and DAMamba-L (${\sim}70$\,M) are evaluated under identical splits, augmentation, loss, optimizer, scheduler, and epoch budget. SegMamba is the fixed-scan ToM reference. In ablation studies, we vary block placement, loss, or scan order on the full-volume pipeline.

Table~\ref{tab:main} summarizes five-fold full-volume performance of DAMamba-UNet3D. Mean Dice is $0.815 \pm 0.013$, with strong WT ($0.897 \pm 0.012$) and TC ($0.826 \pm 0.018$) scores. ET remains hardest ($0.721 \pm 0.014$) due to small lesion volume. Surface distances are favorable for WT and TC (mean HD95 5.9\,mm and 8.4\,mm).\label{sec:main_results}

\begin{table}[t]
\centering
\caption{Full-volume 5-fold cross-validation on BraTS~2020 (DAMamba-UNet3D: DAS at E2--E4, ${\sim}5.3$\,M parameters). Per-case Dice as per the BraTS-official convention.}
\label{tab:main}
\adjustbox{max width=\textwidth,center}{%
\footnotesize\setlength{\tabcolsep}{3pt}%
\begin{tabular}{l|c|c|c|c|c|c|c}
\toprule
\textbf{Fold} & \textbf{Mean} & \textbf{ET} & \textbf{TC} & \textbf{WT} & \textbf{HD95 ET} & \textbf{HD95 TC} & \textbf{HD95 WT} \\
\midrule
0 & 0.824 & 0.740 & 0.829 & 0.903 & 30.0 & 6.8 & 4.6 \\
1 & 0.835 & 0.733 & 0.859 & 0.912 & 20.8 & 4.7 & 4.9 \\
2 & 0.801 & 0.718 & 0.806 & 0.879 & 34.9 & 12.3 & 6.8 \\
3 & 0.814 & 0.717 & 0.823 & 0.903 & 36.3 & 6.9 & 6.3 \\
4 & 0.800 & 0.699 & 0.814 & 0.888 & 36.5 & 11.6 & 6.8 \\
\midrule
\textbf{Mean$\,\pm\,$Std} & \textbf{0.815$\,\pm\,$0.013} & \textbf{0.721$\,\pm\,$0.014} & \textbf{0.826$\,\pm\,$0.018} & \textbf{0.897$\,\pm\,$0.012} & \textbf{31.7} & \textbf{8.4} & \textbf{5.9} \\
\bottomrule
\end{tabular}}%
\end{table}
Table~\ref{tab:comparison} compares segmentation performance by the DAMamba-UNet3D, SegMamba, and DAMamba-L. DAMamba-UNet3D delivers near-parity at ${\sim}5.3$\,M ($0.815 \pm 0.013$ vs.\ SegMamba $0.824 \pm 0.014$). DAMamba-L surpasses SegMamba at large scale ($0.829 \pm 0.012$ vs.\ $0.824 \pm 0.014$; +0.5\,pt mean Dice), with the largest gain on TC ($0.842$ vs.\ $0.827$). Because DAMamba-L differs from SegMamba in U-Net depth, block stacking, and bottleneck design as well as scan type, we attribute this improvement to the full encoder-only DAS-native hybrid rather than to scan protocol alone; nevertheless, the result indicates that learned tri-plane DAS can exceed fixed ToM under matched training and full-volume evaluation on BraTS~2020. The compact ${\sim}0.009$ gap of DAMamba-UNet3D reflects lower parameter count and shallower encoder stacking, not a fundamental limit of adaptive scanning.\label{sec:segmamba}

\begin{table}[t]
\centering
\caption{Accuracy--parameter comparison on BraTS~2020 (full-volume per-case Dice; identical protocol). DAMamba-L uses encoder-only DAMamba (E2--E4) and a convolutional bottleneck.}
\label{tab:comparison}
\adjustbox{max width=\textwidth,center}{%
\footnotesize\setlength{\tabcolsep}{3pt}%
\begin{tabular}{@{}p{0.34\linewidth}r|c|c|c|c@{}}
\toprule
\textbf{Method} & \textbf{Params} & \textbf{Mean} & \textbf{ET} & \textbf{TC} & \textbf{WT} \\
\midrule
\textbf{DAMamba-L} & ${\sim}$70M & \textbf{0.829$\,\pm\,$0.012} & \textbf{0.745$\,\pm\,$0.013} & \textbf{0.842$\,\pm\,$0.014} & 0.901$\,\pm\,$0.013 \\
SegMamba~\cite{ma2024segmamba} (retrained) & ${\sim}$70M & 0.824$\,\pm\,$0.014 & 0.744$\,\pm\,$0.009 & 0.827$\,\pm\,$0.023 & 0.901$\,\pm\,$0.015 \\
\textbf{DAMamba-UNet3D} & \textbf{5.3M} & 0.815$\,\pm\,$0.013 & 0.721$\,\pm\,$0.014 & 0.826$\,\pm\,$0.018 & 0.897$\,\pm\,$0.012 \\
\bottomrule
\end{tabular}}%
\end{table}

\subsection{Component Ablations}
\label{sec:ablation}

All ablations use five folds, 50 epochs, and the full-volume evaluator (Sec.~\ref{sec:eval}). Table~\ref{tab:ablation} reports mean Dice across folds. Our model places DAMamba only at E2--E4 (0.815), outperforming the E2--E4 + bottleneck variant (0.807) and the convolution-only baseline (0.808). Decoder DAMamba further lowers Dice to 0.805. Loss ablations on the bottleneck architecture confirm that the modulated combo loss clearly outperforms CE-only (0.798).

\begin{table}[t]
\centering
\caption{Component ablations (full-volume per-case Dice, 5-fold mean~$\pm$~std; all rows use modulated combo loss unless noted).}
\label{tab:ablation}
\adjustbox{max width=\textwidth,center}{%
\footnotesize\setlength{\tabcolsep}{4pt}%
\begin{tabular}{@{}p{0.65\linewidth}r@{}}
\toprule
\textbf{Configuration} & \textbf{Mean Dice} \\
\midrule
Baseline 3D U-Net (no DAMamba) & 0.808$\,\pm\,$0.012 \\
E2--E4 + bottleneck DAS & 0.807$\,\pm\,$0.012 \\
Full DAMamba (+ decoder blocks) & 0.805$\,\pm\,$0.012 \\
Loss: cross-entropy only\textsuperscript{$\dagger$} & 0.798$\,\pm\,$0.014 \\
Loss: CE + Dice, unmodulated\textsuperscript{$\dagger$} & 0.806$\,\pm\,$0.012 \\
\midrule
\textbf{DAMamba-UNet3D (E2--E4, modulated combo)} & \textbf{0.815$\,\pm\,$0.013} \\
\bottomrule
\multicolumn{2}{@{}p{0.95\linewidth}@{}}{\textsuperscript{$\dagger$}Loss ablations use E2--E4}
\end{tabular}}%
\end{table}

\subsection{Scan-Pattern Ablations}
\label{sec:scan_ablation}

To compare scan orders, we report fixed raster, snake, cross, and tri-orientated (SegMamba-style) flattening under an auxiliary E2--E4 layout (Table~\ref{tab:scan_ablation}). The fair learned-DAS baseline is our \emph{final} model (E2--E4 only, Table~\ref{tab:main}): removing bottleneck DAMamba raises learned DAS from 0.807 to 0.815 (+0.8\,pt), exceeding the best fixed order in the auxiliary study (tri-orientated, 0.808) by 0.7\,pt. 

\begin{table}[t]
\centering
\caption{Scan-pattern ablations (full-volume per-case Dice, 5-fold mean~$\pm$~std). Final model uses learned DAS at E2--E4 only (Table~\ref{tab:main}); fixed orders use auxiliary E2--E4 + bottleneck layout.}
\label{tab:scan_ablation}
\adjustbox{max width=\textwidth,center}{%
\footnotesize\setlength{\tabcolsep}{4pt}%
\begin{tabular}{@{}p{0.62\linewidth}r@{}}
\toprule
\textbf{Scan order / placement} & \textbf{Mean Dice} \\
\midrule
\multicolumn{2}{@{}l@{}}{Fixed orders (E2--E4):} \\
~~~Raster & 0.805$\,\pm\,$0.013 \\
~~~Snake & 0.807$\,\pm\,$0.011 \\
~~~Cross & 0.806$\,\pm\,$0.012 \\
~~~Tri-orientated & 0.808$\,\pm\,$0.010 \\
Learned DAS (E2--E4 + bottleneck) & 0.807$\,\pm\,$0.012 \\
\textbf{Learned DAS (E2--E4; final model)} & \textbf{0.815$\,\pm\,$0.013} \\
\bottomrule
\end{tabular}}%
\end{table}

\subsection{Qualitative Results}

\begin{figure}[t]
\centering
\includegraphics[width=10cm]{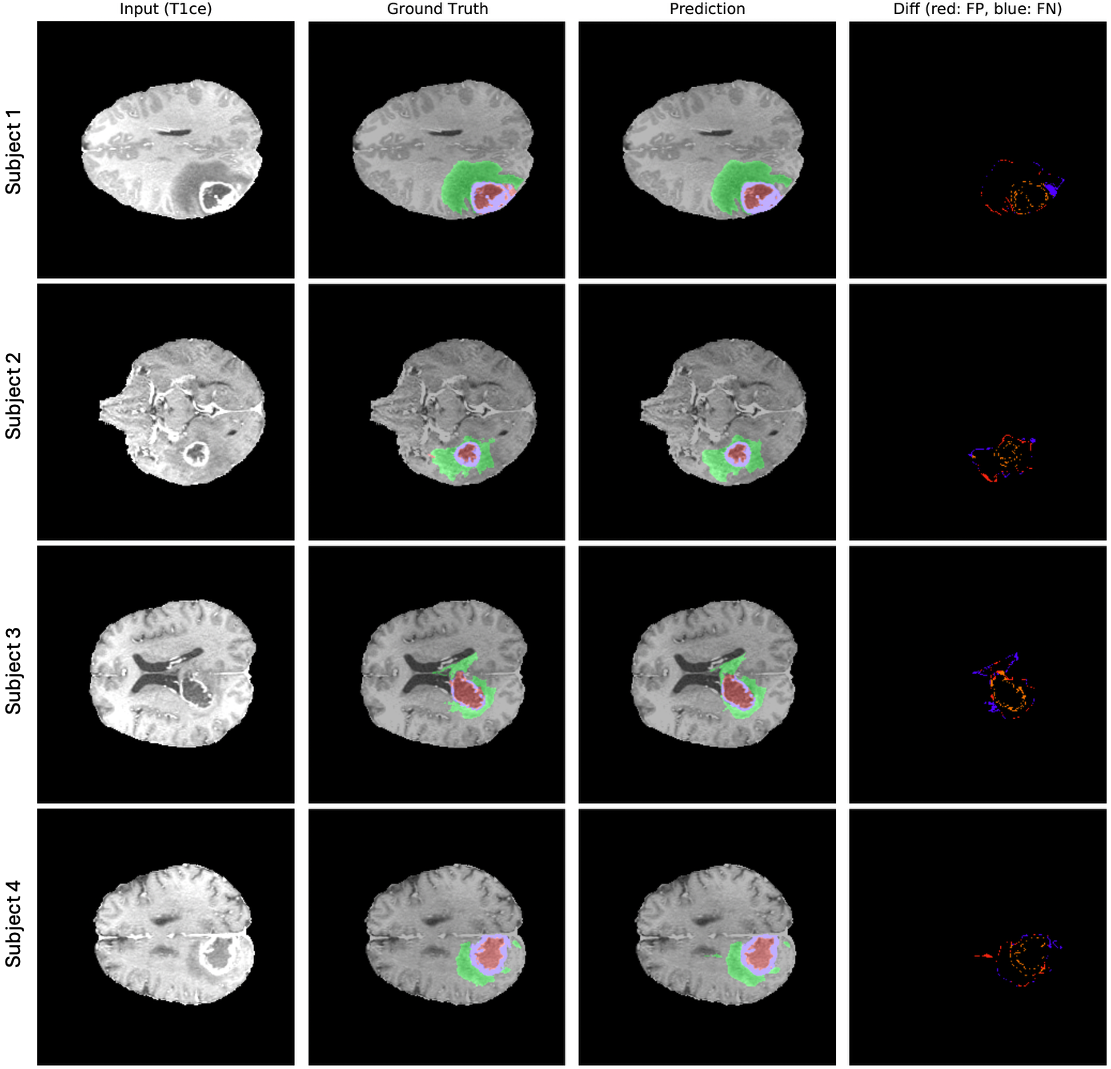}
\caption{Qualitative full-volume predictions on BraTS~2020 validation subjects.}
\label{fig:qualitative}
\end{figure}

Figure~\ref{fig:qualitative} shows qualitative segmentation results on validation data. We display four randomly selected subjects (one per row), each at three axial depths that contain tumor. Each row shows: (a) input T1ce, (b) ground truth (red: necrosis/NCR, green: edema/ED, pink: enhancing tumor/ET), (c) prediction, and (d) difference map (errors only: red=FP, blue=FN). All four subjects demonstrate good segmentation performance, with the model capturing tumor boundaries across varying depths.


\section{Discussion}
\label{sec:discussion}

\noindent\textbf{Two-scale results.} Table~\ref{tab:comparison} supports a dual narrative. DAMamba-UNet3D (${\sim}5.3$\,M; $0.815$ mean Dice) targets parameter-efficient deployment at ${\sim}13\times$ lower capacity than SegMamba with near-parity. DAMamba-L (${\sim}70$\,M; $0.829$) surpasses retrained SegMamba ($0.824$) on the same BraTS~2020 protocol, with the clearest improvement on TC (+1.5\,pt).

\noindent\textbf{Encoder-only placement.} Component ablations on the compact model show that DAMamba at the bottleneck or decoder lowers Dice (Table~\ref{tab:ablation}). DAMamba-L adopts the same encoder-only, convolutional-bottleneck design at SegMamba scale ($0.829$ mean Dice). Placement therefore matters as much as scan type.

\noindent\textbf{Learned DAS vs.\ fixed ToM.} Scan ablations on the compact model show learned DAS ahead of fixed orders when encoder placement is matched (Table~\ref{tab:scan_ablation}). At SegMamba scale, the full DAS-native hybrid exceeds fixed ToM SegMamba. We do not isolate scan protocol from architecture in a single swap experiment, so we state the inference cautiously: on BraTS~2020, learned tri-plane DAS in our hybrid U-Net \emph{appears competitive with and may improve upon} fixed ToM under identical training and full-volume evaluation.

\noindent\textbf{Limitations.} All experiments use BraTS~2020 only. Despite DAS enabling performance in parity with fewer parameter counts than its fixed-scan-based counterpart, multi-dataset validation and statistical testing are needed before general claims about DAS versus ToM.

\noindent\textbf{Broader impact.} Regardless of the BraTS scores above, extending DAS to tri-plane 3D volumes and applying it to medical segmentation for the first time is a substantive contribution in its own right. It offers a learnable alternative to fixed scan trajectories that can be reused in other volumetric tasks without redesigning the core operator. In addition, it opens a way to reduce parameter space multi-folds without losing performance significantly.

\section{Conclusion}
\label{sec:conclusion}

We presented DAMamba-UNet3D, the first 3D medical extension of Dynamic Adaptive Scan in a parameter-efficient hybrid U-Net (encoder-only DAMamba at E2--E4, ${\sim}5.3$\,M). On BraTS~2020 it achieves near-parity with retrained SegMamba at ${\sim}13\times$ lower capacity ($0.815 \pm 0.013$ vs.\ $0.824 \pm 0.014$). At large scale, DAMamba-L with encoder-only DAS and a convolutional bottleneck surpasses SegMamba ($0.829 \pm 0.012$ vs.\ $0.824 \pm 0.014$), with component ablations confirming that encoder-only placement is essential. The combined evidence suggests that learned tri-plane DAS in a hybrid U-Net is a strong alternative to fixed ToM scanning for 3D brain tumor segmentation. We will publish our code upon acceptance of this paper.

\bibliographystyle{splncs}
\bibliography{references_EMAI}

\end{document}